\documentclass[letterpaper, 10 pt, conference]{ieeeconf}  

\IEEEoverridecommandlockouts                              

\usepackage{times}
\usepackage{soul}
\usepackage{url}
\usepackage[hidelinks]{hyperref}
\usepackage[utf8]{inputenc}
\usepackage[small]{caption}
\usepackage{graphicx}
\usepackage{amsmath}
\usepackage{booktabs}
\usepackage{array}
\usepackage{algorithm}
\usepackage{algorithmic}
\usepackage[switch]{lineno}
\usepackage{amsfonts}

\title{\LARGE \bf
TactileNet: Bridging the Accessibility Gap with AI-Generated Tactile Graphics for Individuals with Vision Impairment
}

\author{Adnan Khan$^{1,*}$,
Alireza Choubineh$^1$,
Mai A. Shaaban$^{2}$,
Abbas Akkasi$^1$ and
Majid Komeili$^1$%
\thanks{$^{1}$School of Computer Science, Carleton University, Ottawa, Canada}%
\thanks{$^{2}$Mohamed bin Zayed University of Artificial Intelligence, Abu Dhabi, UAE}%
\thanks{$^{*}$Corresponding author: Adnan Khan. E-mail: adnankhan5@cmail.carleton.ca}%
}

\begin{document}

\maketitle
\thispagestyle{empty}
\pagestyle{empty}

\begin{abstract}

Tactile graphics are essential for providing access to visual information for the 43 million people globally living with vision loss. Traditional methods for creating these graphics are labor-intensive and cannot meet growing demand. We introduce \textbf{TactileNet}, the first comprehensive dataset and AI-driven framework for generating embossing-ready 2D tactile templates using text-to-image Stable Diffusion (SD) models. By integrating Low-Rank Adaptation (LoRA) and DreamBooth, our method fine-tunes SD models to produce high-fidelity, guideline-compliant graphics while reducing computational costs. Quantitative evaluations with tactile experts show 92.86\% adherence to accessibility standards. Structural fidelity analysis revealed near-human design similarity, with an SSIM of 0.538 between generated graphics and expert-designed tactile images. Notably, our method preserves object silhouettes better than human designs (SSIM = 0.259 vs. 0.215 for binary masks), addressing a key limitation of manual tactile abstraction. The framework scales to 32,000 images (7,050 high-quality) across 66 classes, with prompt editing enabling customizable outputs (e.g., adding or removing details). By automating the 2D template generation step—compatible with standard embossing workflows—TactileNet accelerates production while preserving design flexibility. This work demonstrates how AI can augment (not replace) human expertise to bridge the accessibility gap in education and beyond. Code, data, and models will be publicly released to foster further research.

\end{abstract}

\section{INTRODUCTION}

Providing accessible alternatives to visual information for individuals with visual impairments is an urgent societal challenge in the digital age. According to the International Agency for the Prevention of Blindness (IAPB), an estimated 1.1 billion people worldwide were living with vision loss in 2020, a number projected to increase by 55\% by 2050 due to population growth and aging \cite{IAPB2020}. Among these, 43 million people are blind, and 295 million have moderate to severe visual impairment, underscoring the urgent need for inclusive solutions. Many individuals with vision loss face significant barriers in education and daily life, as learning materials and information systems remain heavily reliant on visual content. This growing disparity highlights the necessity for scalable, innovative approaches to improve accessibility and bridge the gap for those with visual impairments.

\begin{figure}[ht]
    \centering
    \begin{tabular}{cc}
    \includegraphics[width=0.43\columnwidth]{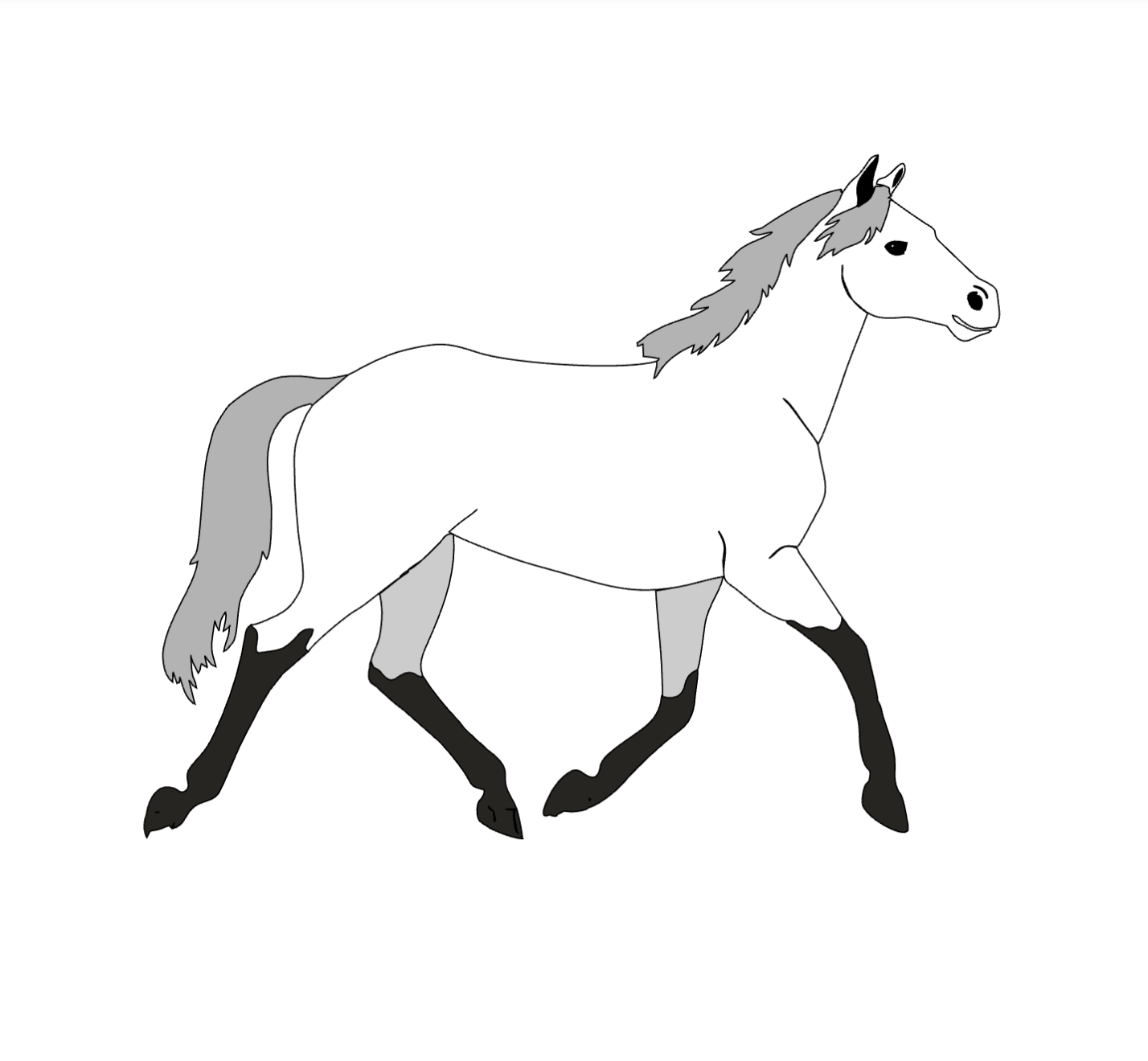} &
    \includegraphics[width=0.43\columnwidth]{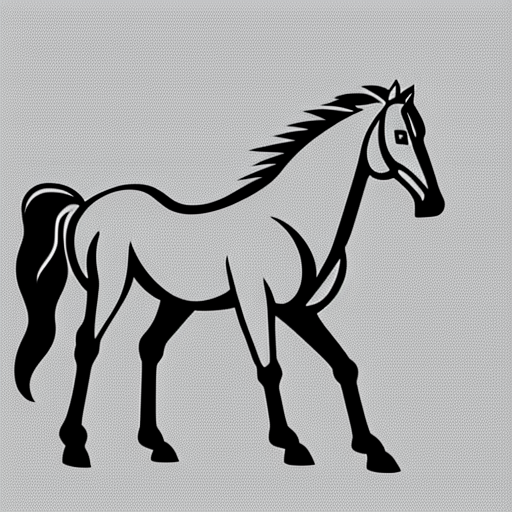} \\
    (a) Sourced tactile graphic & (b) Generated tactile graphic
    \end{tabular}
    \caption{Example tactile graphics of a horse. Left: Expert-designed tactile graphic from existing libraries. Right: Tactile graphic generated by our fine-tuned model.}
    \label{fig:intro_horse}
\end{figure}

Tactile graphics—raised, textured representations of visual content—serve as a vital medium for conveying graphical information through touch, adhering to strict guidelines (e.g., line thickness, spacing) set by the Braille Authority of North America (BANA) \cite{BrailleAuthority} to ensure clarity and usability. Figure~\ref{fig:intro_horse} illustrates an example of such graphics, highlighting how both expert-designed and model-generated images must differentiate between near-side and far-side limbs to convey posture and depth—an essential consideration in tactile design for side-view animal depictions. Despite advancements in design tools (e.g., CorelDRAW, Adobe Illustrator, TactileView) and embossing technologies (e.g., ViewPlus and Index Braille), manual tactile graphic creation remains prohibitively labor-intensive, with simple images requiring 10–15 minutes and complex images taking many hours to produce \cite{washingtonTactileGraphics,APHKit}. Existing AI-driven solutions \cite{mukhiddinov2021systematic} are hindered by the absence of large-scale, domain-specific datasets, forcing reliance on synthetic data that fails to capture tactile design nuances.
Consequently, these technologies have yet to be widely integrated into educational systems.

To address these scalability challenges, we introduce TactileNet, the first comprehensive digital dataset designed to train adapters for AI models that generate tactile images from text prompts or combined text and natural image inputs. Given the absence of paired datasets, these models are initially trained using text-to-image Stable Diffusion (SD) models \cite{ho2020denoising,rombach2021highresolution} and later evaluated under image-to-image translation scenarios to ensure practical usability. This step is crucial for preserving essential structural information while omitting extraneous details like color and complex textures, which may hinder tactile perception.

\begin{figure*}[t!]
\centering
\includegraphics[width=\textwidth]{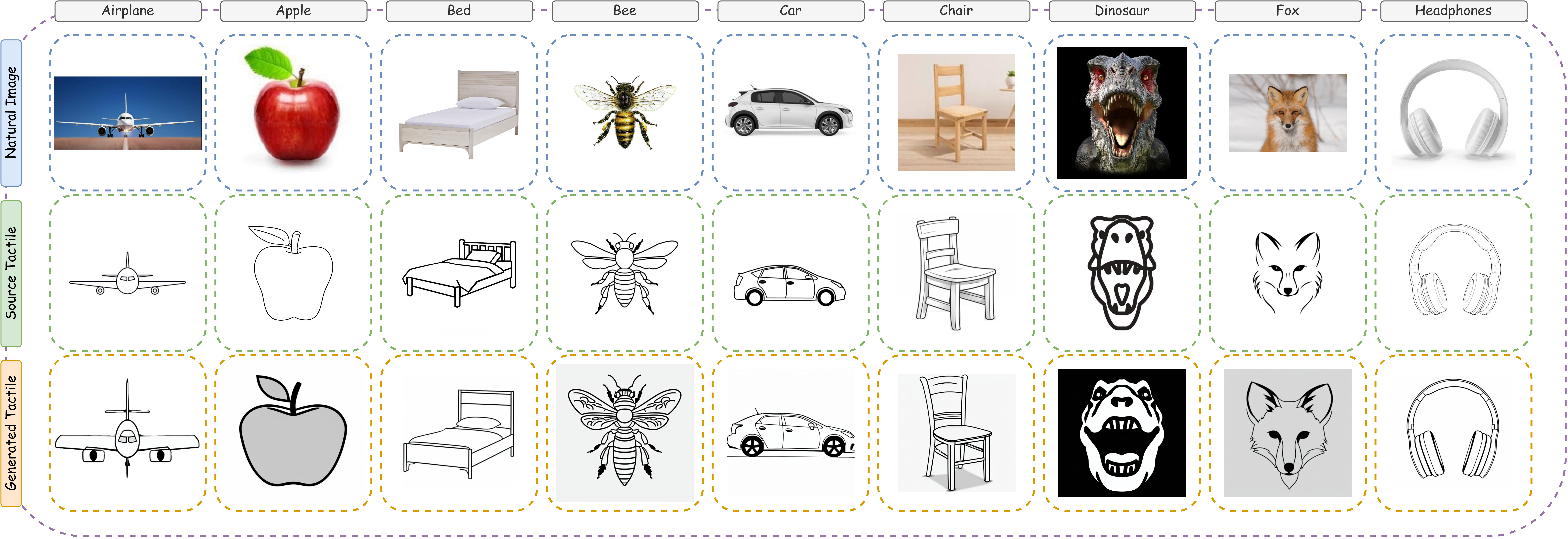}
\caption{Examples of our image-to-image translation framework: Top row shows reference natural images, middle row displays TactileNet's sourced/benchmark tactile graphics, and the bottom row presents generated tactile graphics using our adapters.}
\label{fig:main}
\end{figure*}
Our framework uniquely integrates Low-Rank Adaptation (LoRA) \cite{hu2021lora} and DreamBooth \cite{Ruiz_2023_CVPR}, enabling parameter-efficient fine-tuning on TactileNet while preserving the base model’s generative diversity—critical for handling rare objects. LoRA reduces computational costs by adapting low-rank parameters, while Dreambooth enables personalized training with minimal data. The model's ability to transform RGB images into tactile formats is validated through human expert feedback, demonstrating its potential to improve tactile graphic quality and accessibility. By automating and streamlining the creation process, our method allows designers to focus on refinement and customization, significantly reducing time and effort. The contributions of this paper are threefold:
\begin{itemize}
\item Introducing TactileNet, a consolidated dataset for training AI models tailored for tactile applications.
\item Developing a novel methodology leveraging SD, LoRA, and Dreambooth to automate tactile graphic generation and text-based editing, aiding tactile designers.
\item Implementing a web-based evaluation protocol to assess model effectiveness through expert human feedback.
\end{itemize}

\section{RELATED WORK}
\label{sec:lit_review}

\subsection{Tactile Graphics: Current Approaches and Limitations}
Recent advances in tactile graphic production have prioritized accessibility for blind and visually impaired (BVI) communities. Early work focused on manual design principles, such as \cite{gonzalez2019tactiled}'s framework for assessing image complexity to determine tactile suitability. While foundational, these methods rely heavily on human expertise—a bottleneck for scalability. Emerging tools like tactile graphics finders \cite{felipe2020ml} and audio-tactile tablets \cite{guinness2019robographics} enhance content discovery but fail to automate the end-to-end design pipeline. Similarly, video description systems \cite{yuksel2020human} improve contextual awareness but lack tactile translation capabilities, leaving a critical gap in accessible content creation.

\subsection{GenAI and Deep Learning in Accessibility}
Generative AI (GenAI) has advanced accessibility through applications like real-time captioning \cite{khan2024improving} and scene segmentation \cite{kirillov2023segment}. However, tactile graphic generation poses unique challenges, as noted in \cite{mukhiddinov2021systematic}'s systematic review: (1) scarcity of paired datasets (text/image → tactile graphics), and (2) high costs of refreshable tactile displays. Text-to-image models like Stable Diffusion (SD) \cite{rombach2021highresolution} offer promise but require adaptation to tactile constraints (e.g., suppressing textures, emphasizing edges). Recent parameter-efficient tuning methods like Low-Rank Adaptation (LoRA) \cite{hu2021lora} address data scarcity but remain untested in tactile domains.

\subsection{Gaps and Contributions}
Existing solutions address fragments of the tactile design workflow but lack holistic automation. Our work bridges three key gaps:
\begin{itemize}
\item \textbf{Dataset Scarcity}: We introduce TactileNet, the first dataset curated for text/image-to-tactile translation.
\item \textbf{Structural Adaptation}: While SD generates photorealistic images \cite{ho2020denoising}, we retrain it to prioritize tactile usability (simplified shapes, high-contrast edges).
\item \textbf{Customization}: Prior tools focus on static outputs; our framework enables prompt-based edits (e.g., \textit{``Remove background clutter''}).
\end{itemize}
This establishes a new benchmark for scalable, AI-assisted tactile design.

\section{METHODOLOGY}
\label{sec:methodology}
This section covers TactileNet dataset curation and SD model adaptation for tactile graphic generation.

\subsection{Preliminaries}
\label{related_works}

\subsubsection{Denoising Diffusion Probabilistic Models (DDPMs)} 
A DDPM \cite{sohldickstein2015deep,ho2020denoising} consists of two Markov chains: a forward chain that gradually adds noise to data, transforming it into a Gaussian distribution, and a reverse chain that learns to denoise and reconstruct the data \cite{koller2009probabilistic}.
\begin{equation}
x_t = \sqrt{\bar{\alpha}_t} x_0 + \sqrt{1-\bar{\alpha}_t} \epsilon, \quad \epsilon \sim \mathcal{N}(0, I)
\end{equation}

Here, $\mathbf{x}_0$ represents the original data, $\bar{\alpha}_t$ controls the noise level at step $t$, and $\boldsymbol{\epsilon}$ is Gaussian noise. The reverse process, also referred to as denoising, reconstructs the original data by progressively estimating and removing the added noise:
\begin{equation}
x_{t-1}=\frac{1}{\sqrt{\alpha_t}}\left(x_t-\frac{1-\alpha_t}{\sqrt{1-\bar{\alpha}_t}} * \epsilon_\theta(x_t, t)\right)
\end{equation}
In this equation, $\boldsymbol{\epsilon}_\theta(\mathbf{x}_t,t)$ is the noise component predicted by the neural network model. 

\subsubsection{Text-to-Image SD Models} 
Text-to-Image SD models extend DDPMs, using textual prompts to guide image generation and enable text-to-visual synthesis. These models represent a convergence of vision-language technologies, where the structured approach of DDPMs is adapted to understand and generate images based on textual cues \cite{radford2021learning}. This involves not only transforming noise into structured images but also integrating linguistic elements to ensure that the generated visuals accurately reflect the described scenarios \cite{yang2024diffusion,kawar2023imagic,du2022survey}.

\subsubsection{Fine-Tuning SD Models}
Fine-tuning adapts pre-trained models to specific tasks, enhancing performance in applications requiring precision and detail. Our framework leverages two methods: LoRA \cite{hu2021lora} and Dreambooth \cite{Ruiz_2023_CVPR}, fine-tuned using Kohya's Trainer \cite{kohyaTrainer2023} on text-image pairs to refine the model’s ability to interpret textual prompts. During inference, the system supports both text-to-image and image-to-image generation (e.g., via the SD web interface \cite{automatic1111}), enabling tactile graphic synthesis from textual descriptions or reference images. This flexibility ensures practical usability.

\subsubsection{Low-Rank Adaptation (LoRA)} 
LoRA is a parameter-efficient fine-tuning method, ideal for resource-constrained scenarios or small datasets. LoRA updates only a subset of the model's parameters, enabling adaptation to new tasks without comprehensive re-training. It introduces a pair of low-rank matrices \(A\) and \(B\) for each weight matrix \(W\). For a weight matrix $W \in \mathbb{R}^{m \times n}$, LoRA introduces $A \in \mathbb{R}^{m \times r}$ and $B \in \mathbb{R}^{r \times n}$, where $r \ll m, n$. The adapted weight matrix becomes $W + \Delta W$, where $\Delta W = AB$. This significantly reduces the number of trainable parameters. The original model parameters remain frozen.

\subsubsection{DreamBooth} 
DreamBooth fine-tunes SD models using a small set of subject images, enabling the generation of customized images with high fidelity. It involves selecting a unique identifier for a specific subject, training the model to associate the subject's characteristics with this identifier. This adjusts the model's parameters to generalize the appearance of the new subject across diverse scenarios.

\subsection{Dataset Creation}
\begin{figure*}[htpt!]
\centering
\includegraphics[width=0.8\textwidth]{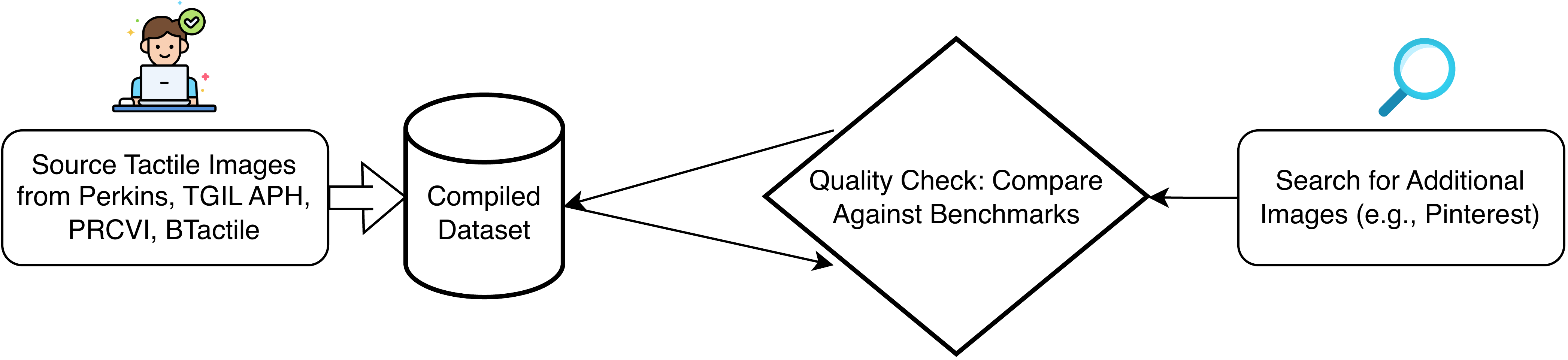} 
\caption{Flow diagram illustrating the process of data compilation from initial sourcing to final dataset compilation.}
\label{fig:data-compilation}
\end{figure*}

\subsubsection{Data Collection}
We curated the TactileNet dataset by sourcing tactile images from online digital libraries, including Perkins College for the Blind, the Tactile Graphics Image Library of American Printing House (TGIL APH), the Provincial Resource Centre for the Visually Impaired (PRCVI), and BTactile \cite{PerkinsTactileGraphicsLibrary,aph,prcvi,BTactile}. These libraries are known for high-quality, expert-designed tactile images.

Given the limited number of available images for certain classes (e.g., only four "dog" images in TGIL APH), we expanded the dataset by adding visually similar images from online platforms like Pinterest. Each image was carefully compared against gold-standard tactile designs to ensure consistency in quality and educational value. Figure~\ref{fig:data-compilation} illustrates the data sourcing process. The final dataset comprises 1,029 tactile images across 66 classes. Table~\ref{tab:data-summary} provides class distribution statistics.

\paragraph{Collaboration with Industry Partners}
Throughout dataset development, we collaborated with industry partners specializing in accessibility solutions, including tactile graphics designers and educators. Their feedback ensured that our dataset met both technical standards and practical requirements for real-world educational use.

\begin{figure*}[t!]
    \centering
    \includegraphics[width=\textwidth]{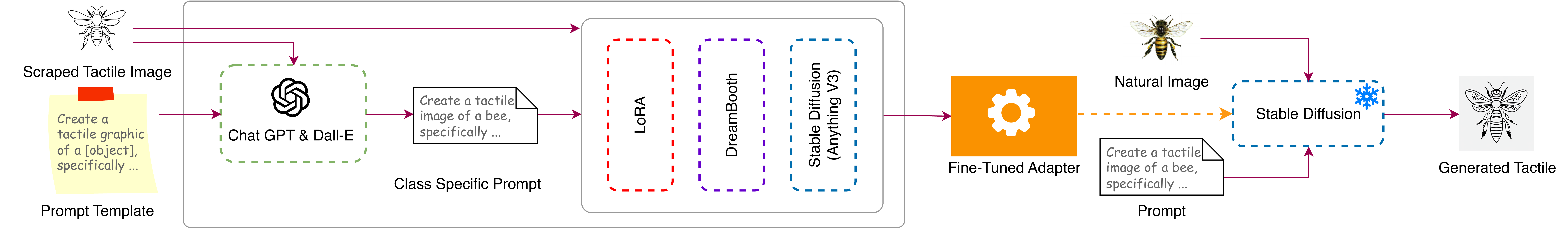}
    \caption{Comprehensive workflow of our framework, starting with fine-tuning (left), where TactileNet data (tactile images, text prompts) refine the SD model. The process transitions to the generation phase (right), applying fine-tuned adapters atop the frozen SD model for text-to-image and image-to-image tactile graphic generation.}
    \label{fig:tactilenet_process}
\end{figure*}

\subsubsection{Text Prompts Generation for Tactile Graphics}
\label{prompt_generation}
We generated text prompts for each tactile image using ChatGPT \cite{chatgpt-dalle} integrated with DALL-E, following a structured template:

\textit{Create a tactile graphic of an [object], specifically designed for individuals with visual impairments. The graphic should feature raised, smooth lines to delineate the [patterns], against a simplistic background.}

The [object] was replaced by the class name (e.g., cat) and [patterns] by specific features (e.g., whiskers, eyes, paws). All prompts were reviewed to ensure adherence to the template and alignment with the tactile graphic.

\subsection{Models Development and Image Generation}

\subsubsection{Fine-Tuning Individual Models for Each Category}
We fine-tuned 66 distinct models (one per class) using LoRA and DreamBooth (Figure~\ref{fig:tactilenet_process}, left). The ``tactile'' identifier ensured alignment with tactile features during training. Implemented via the Kohya Trainer \cite{kohyaTrainer2023}, this process adapted the Stable Diffusion-based Anything V3 model \cite{anythingv3,rombach2021highresolution}. Anything V3 was selected after comparing generation quality across SD v1 and v1.5. The fine-tuned adapter yielded superior results when used atop the base SD v1.5 model.

\subsubsection{Generation of Tactile Images}
Following fine-tuning, tactile graphics were generated using the fine-tuned SD v1.5 model (Figure~\ref{fig:tactilenet_process}, right). Two generation settings were used: (1) class-specific text prompts alone and (2) prompts combined with natural images. Including natural images doubled the output quality, with 30\% of graphics meeting standards versus 15\% for text-only prompts (Section~\ref{evaluation_protocol}).

\section{EXPERIMENTAL SETTINGS}
\label{sec:experimental_settings}

This section outlines the configuration and environmental settings utilized in our experiments.

\subsection{Baseline: Instruction-Tuned Stable Diffusion (InstructPix2Pix)}

As a baseline, we fine-tuned Stable Diffusion (SD) \cite{rombach2021highresolution} using the InstructPix2Pix method \cite{Paul2023instruction-tuning-sd,brooks2023instructpix2pixlearningfollowimage} on 1,029 triplets consisting of natural images, sourced tactile graphics, and corresponding edit prompts. The natural images were gathered by a team of five undergraduate students, who reviewed each sourced tactile graphic and selected the best-matching natural image from online sources. 

The edit prompts were deliberately kept simple and generic—e.g., \textit{``Create a tactile image of the given natural image of a [class].''}—to align with the original instruction-tuned training paradigm. Over 100 such generic prompts were created using ChatGPT and randomly paired with images, with [class] replaced by the object name. We also experimented with complex, template-based prompts detailing features and patterns. In both cases, however, the model exhibited significant overfitting: during testing, it disregarded input images and memorized the prompt-class pairs from the training data. Changing prompts often resulted in hallucinated or irrelevant outputs, highlighting the baseline’s strong dependence on text rather than visual conditioning.

\begin{figure}[ht]
    \centering
\includegraphics[width=0.65\columnwidth]{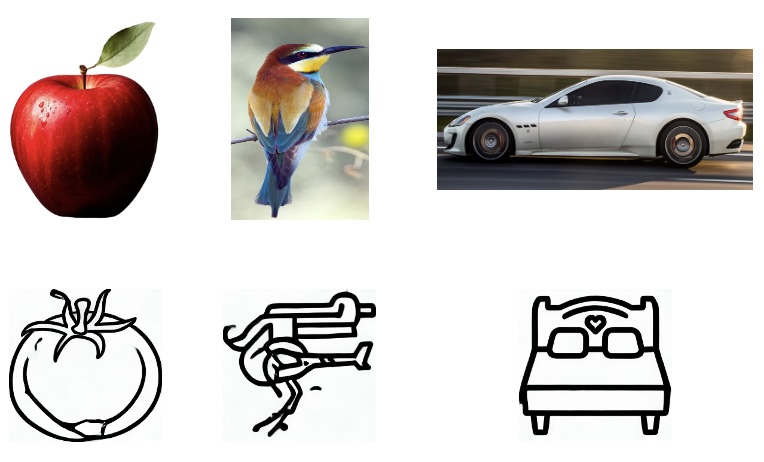}
    \caption{Failure cases from the instruction-tuned baseline. Although the test-time natural images (top row) are diverse and unseen during training, the model fails to adapt and instead memorizes training prompt-class pairs, producing incorrect tactile outputs (bottom row). Memorized outputs correspond to tomato, helicopter, and bed classes.}
\label{fig:baseline_overfitting}
\end{figure}

To increase data diversity and mitigate overfitting, we expanded the dataset by pairing each of the 1,029 tactile-natural image pairs with the full set of 100 generic prompts, resulting in 102,900 training examples. Nevertheless, the baseline underperformed on unseen test images, defaulting to memorized prompt outputs rather than adapting to new inputs (see Figure~\ref{fig:baseline_overfitting}). Due to severe overfitting and failure to generate usable outputs, the baseline could not achieve measurable Q1/Q2 alignment or SSIM scores. As a result, quantitative comparison metrics are not reported.
These findings highlight the challenge of applying instruction-tuned diffusion models directly for tactile graphic translation tasks, particularly when the model fails to balance prompt guidance with visual conditioning. Moreover, the relatively low pixel and color complexity of tactile images makes them poorly suited for standard full-model fine-tuning approaches, which typically require large, information-rich datasets to avoid overfitting. Our approach using LoRA and Dreambooth better mitigates these challenges by enabling more effective and data-efficient adaptation.
\subsection{Fine-Tuning Configuration}
\label{subsec:model_specs}
We fine-tuned Anything V3 \cite{anythingv3} across 66 distinct classes. Each class had between 9 (llama) and 102 (helicopter) training images, averaging approximately 16 images per class. We leveraged sourced tactile images along with the prompts to develop class-specific adapters. These lightweight modules specialize in generating tactile graphics for each category and are integrated atop frozen versions of SD v1.5 for image generation.
For fine-tuning, the Dreambooth configuration prioritized prior preservation with a loss weight of 1.0, ensuring the model retained its ability to generate generic images while adapting to tactile-specific features. The LoRA setup used a network module with linear dimensions set to 32 and an alpha parameter of 16, controlling the scaling of low-rank updates. This configuration enabled effective parameter adaptation without requiring pre-loaded weights.

\subsubsection{Optimization and Hardware Configuration Parameters}
\label{subsubsec:additional_params}

The optimization process employed the AdamW8bit optimizer \cite{kingma2017adam}, with a learning rate of $1 \times 10^{-4}$ for the UNet \cite{ronneberger2015u} and $5 \times 10^{-5}$ for the text encoder, using a constant rate scheduler without warm-up steps. Fine-tuning experiments were facilitated by NVIDIA Tesla T4 GPUs via Google Colab. Training used a batch size of 6 with mixed precision (FP16) across up to 20 training epochs.

\subsection{Image Generation Configuration}
\label{subsec:image_gen_config}

Text-to-image generation employed the DPM++ 2M Karras sampling method, using 20 steps at 512×512 pixel resolution and a CFG scale of 7. For image-to-image translation, configurations were similar but used a denoising strength of 0.9 to retain the structural essence of the original images while incorporating tactile features.

For challenging classes, deviations were necessary to optimize structure and clarity. Objects requiring finer details—such as bee wings, elephant textures, and duck feathers—used higher denoising strengths (0.96–1.0) and adjusted CFG scales (8–10) for improved edge clarity. Classes with intricate internal features (e.g., basketball, dinosaur teeth, camera lenses) leveraged ControlNet lineart modules for contour refinement. Water-based objects (e.g., sailboat, floating duck) employed negative prompts to suppress extraneous water details. Simpler structures (e.g., egg, book, hatchback car) used LoRA-only adaptations without ControlNet assistance. Image generation was conducted on NVIDIA TITAN V GPUs (12288 MiB memory), with an average generation time of approximately 10 seconds per image.

\subsection{Evaluation Protocol}
\label{evaluation_protocol}

To assess the quality and usability of the generated tactile graphics, we developed a custom evaluation protocol in collaboration with tactile graphics designers. Conventional metrics such as Fréchet Inception Distance (FID) \cite{heusel2017gans} and Structural Similarity Index (SSIM) \cite{wang2004image} do not fully capture tactile-specific requirements like line clarity, texture, and adherence to accessibility guidelines \cite{BrailleAuthority}. Therefore, we prioritized expert evaluation by our industry partners, while also employing SSIM as a supplementary quantitative measure of structural fidelity.

\subsubsection{Interface Design and Rationale}

Our evaluation interface presented evaluators with pairs of images:
\begin{itemize}
    \item \textbf{Reference Image}: A natural image depicting the subject (e.g., a cat standing).
    \item \textbf{Tactile Image}: Either a generated tactile graphic or a sourced tactile graphic from established libraries (e.g., APH, Perkins).
\end{itemize}
This side-by-side comparison allowed evaluators to assess how well the tactile graphic captured essential features and posture while adhering to tactile design principles. Evaluators were blind to the graphic's source (generated vs. sourced), ensuring unbiased assessment. For evaluation, each class included one reference natural image and two tactile graphics (one generated, one sourced), totaling 66 natural images and 132 tactile graphics.

\subsubsection{Evaluation Questions and Metrics}
Evaluators are asked to answer the following questions for each tactile graphic:

\paragraph{Q1: Natural Features and Posture Alignment.}
Determine whether the tactile image accurately reflects the natural features and pose depicted in the reference image. \\
\textbf{Example:} If the reference image shows a cat standing, but the tactile image depicts a cat sitting, this should be marked as 'No'. \\
\textbf{Instructions:} Select 'Yes' if the tactile image aligns well with the reference image in terms of pose and essential features (e.g., body posture, visible organs). Select 'No' if there are discrepancies.

\paragraph{Q2: Adherence to Tactile Graphics Guidelines.}
Assess whether the tactile graphic follows established tactile graphics standards (e.g., BANA guidelines \cite{BrailleAuthority}). \\
\textbf{Example:} A tactile graphic with overly complex patterns that might confuse tactile reading should be marked as 'No'. \\
\textbf{Instructions:} Select 'Yes' if the tactile image adheres to the guidelines. Select 'No' if it fails to meet these standards.

\paragraph{Q3: Quality Rating of the Tactile Image.}
Rate the quality of the tactile graphic based on its utility and adherence to tactile representation principles. \\
\textbf{Options:}
\begin{itemize}
    \item \textbf{Accept as Is}: The tactile image meets all quality standards and requires no modifications.
    \item \textbf{Accept with Minor Edits}: The image is generally acceptable but requires minor modifications to enhance clarity or adherence to guidelines.
    \item \textbf{Accept with Major Edits}: The image requires significant changes to be useful as a tactile graphic.
    \item \textbf{Reject (Useless)}: The image does not meet the standards for tactile graphics and cannot be salvaged through edits.
\end{itemize}
\textbf{Instructions:} Choose the option that best describes the state of the tactile graphic.

\paragraph{Q4: Optional Feedback.}
Provide detailed comments or suggestions for improving the tactile graphic. \\
\textbf{Instructions:} Use this section to highlight specific issues (e.g., line clarity, texture) or suggest modifications.

\subsubsection{Natural Image Selection and Prompt Generation}
Natural images were carefully selected to match the sourced gold-standard tactile graphics from established libraries. A team of five undergraduate students was trained to find the closest matches using Google search queries such as "side profile of [class]." Ambiguities were resolved through max voting when multiple candidates were identified.
For each class, two test samples were prepared that would go with the reference natural image:
\begin{itemize}
    \item \textbf{Sample 1}: A generated tactile graphic produced by our class-specific adapters.
    \item \textbf{Sample 2}: A sourced tactile graphic from established libraries or open-source platforms.
\end{itemize}
The generated tactile graphics were produced using two types of prompts:
\begin{itemize}
    \item \textbf{Original Prompt}: The prompt generated during the prompt generation phase (Section~\ref{prompt_generation}).
    \item \textbf{Paraphrased Prompt}: A refined version of the original prompt, paraphrased using DeepSeek \cite{deepseek} with the instruction: "Paraphrase the given prompt for tactile generation while preserving the 'tactile' subject and essential features."
\end{itemize}
For each prompt type, eight tactile graphics were generated, and the best match was selected based on the evaluation criteria as discussed.

\subsubsection{Statistical Analysis Plan}
The evaluation results were analyzed using binary metrics (Yes/No) for Q1 and Q2, and categorical metrics (Accept as Is, Accept with Minor Edits, etc.) for Q3. Percentages and averages were computed to compare the performance of generated vs. sourced tactile graphics. Detailed results and analysis are presented in the following section.

\section{RESULTS AND DISCUSSION}
\label{sec:results&discussion}

\begin{table*}[ht]
\centering
\resizebox{\textwidth}{!}{%
\begin{tabular}{cccccc}
\toprule
\textbf{Class} & \textbf{(Source, Generated) Counts} & \textbf{Class} & \textbf{(Source, Generated) Counts} & \textbf{Class} & \textbf{(Source, Generated) Counts} \\
\midrule
Airplane       & (10, 55)       & Apple          & (11, 28)       & Ball           & (25, 83)       \\
Banana         & (13, 156)      & Bat            & (13, 60)       & Bed            & (14, 95)       \\
Bee            & (13, 61)       & Beluga Whale   & (11, 27)       & Bicycle        & (11, 117)      \\
Bird           & (16, 115)      & Boat           & (18, 19)       & Book           & (12, 73)       \\
Bottle         & (11, 137)      & Camel          & (10, 109)      & Camera         & (12, 219)      \\
Car            & (25, 106)      & Cat            & (22, 142)      & Chair          & (12, 117)      \\
Clover         & (10, 23)       & Crab           & (10, 321)      & Cup            & (15, 380)      \\
Dinosaur       & (20, 184)      & Dog            & (21, 119)      & Door           & (13, 299)      \\
Duck           & (12, 399)      & Egg            & (17, 87)       & Elephant       & (20, 29)       \\
Fish           & (30, 130)      & Flower         & (13, 72)       & Fox            & (13, 163)      \\
Giraffe        & (12, 12)       & Glasses        & (12, 44)       & Guitar         & (18, 70)       \\
Hammer         & (19, 133)      & Hat            & (12, 73)       & Headphones     & (11, 53)       \\
Helicopter     & (102, 35)      & Horse          & (23, 93)       & Hut            & (10, 226)      \\
Iron           & (10, 110)      & Jellyfish      & (9, 21)        & Lamp           & (10, 38)       \\
Laptop         & (11, 153)      & Leaf           & (11, 127)      & Llama          & (10, 137)      \\
Motorcycle     & (14, 161)      & Pencil         & (19, 85)       & Penguin        & (12, 149)      \\
Planet         & (12, 36)       & Rabbit         & (10, 205)      & Ring           & (10, 17)       \\
Rocket         & (23, 84)       & Satellite      & (10, 49)       & School Backpack & (12, 24)      \\
Scooty         & (12, 46)       & Ship           & (13, 13)       & Shirt          & (21, 152)      \\
Shoe           & (18, 207)      & Snowflake      & (11, 56)       & Soda Cans      & (12, 35)       \\
Spoon          & (10, 25)       & Teddy Bear     & (16, 39)       & Train          & (18, 202)      \\
Tree           & (22, 107)      & Watch          & (11, 83)       & Umbrella       & (10, 25)       \\
\midrule
\multicolumn{6}{c}{\textbf{Total} (1029, 7050) 
\textbf{Mean} (15.4, 123.5) 
\textbf{Median} (12, 93) 
\textbf{Max} (102, 399) 
\textbf{Min} (9, 12)} \\
\bottomrule
\end{tabular}%
}
\caption{Overview of the dataset statistics: scrapped tactile images (\textbf{Source}) used for training the models and the output tactile images (\textbf{Generated}) of our text-to-image adapters.}
\label{tab:data-summary}
\end{table*}

In this section, we present results including TactileNet dataset statistics (Table~\ref{tab:data-summary}) and outcomes from text-to-image and image-to-image translation tasks.

\subsection{Image-to-Image Translation Evaluation}  
Table~\ref{tab:quality_ratings} summarizes the quality ratings for both sourced and generated tactile images. Figure~\ref{fig:main} illustrates sample outputs from our evaluation: the first row presents the natural reference images, the second row displays the sourced tactile graphics, and the third row showcases the tactile graphics generated using our fine-tuned adapters.

\begin{table}[ht]
\centering
\caption{Quality Ratings for Generated vs. Sourced Tactile Graphics}
\begin{tabular}{lcc}
\toprule
\textbf{Category} & \textbf{Generated (\%)} & \textbf{Sourced (\%)} \\
\midrule
Accept as Is & 32.14 & \textbf{35.71} \\
Accept with Minor Edits & 39.29 & 39.29 \\
Accept with Major Edits & 28.57 & \textbf{21.43} \\
Reject (Useless) & \textbf{00.00} & 3.57 \\
\bottomrule
\end{tabular}
\label{tab:quality_ratings}
\end{table}  

\begin{figure}[ht]  
\centering  
\includegraphics[width=0.8\columnwidth]{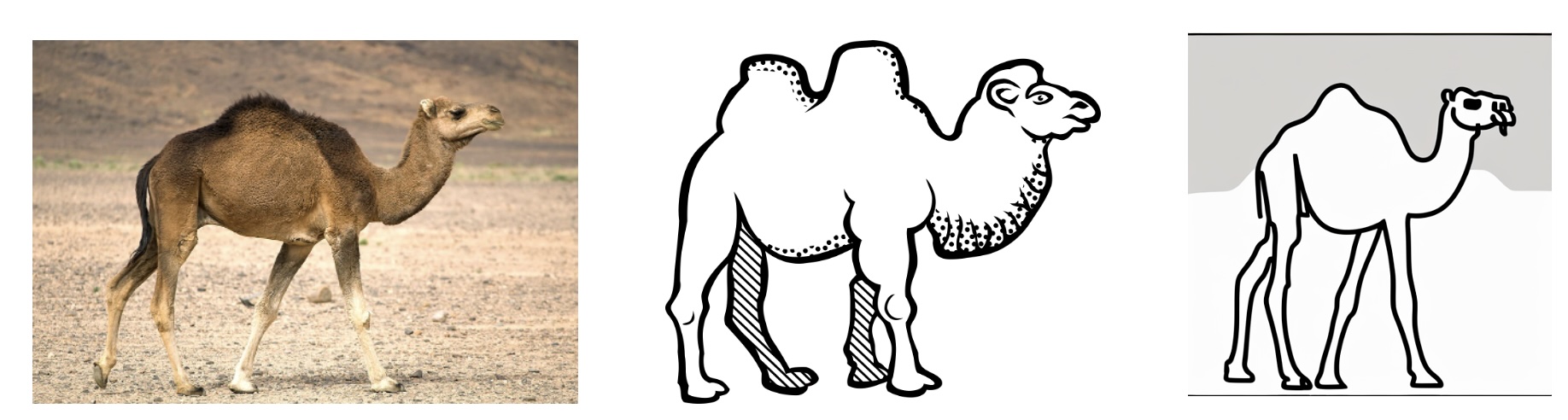}  
\caption{Example of human-induced errors in natural image pairing: (Left) Natural image of a Bactrian camel; (Center) Sourced tactile graphic of a Dromedary camel; (Right) Generated tactile graphic adhering to the reference.}  
\label{fig:camel_mismatch}  
\end{figure}  

\subsubsection{Key Findings} 
Our evaluation reveals that both sourced and generated tactile graphics achieved \textbf{100\% alignment} with natural images in terms of pose and structural features (\textbf{Q1}), confirming their accuracy in representation. However, adherence to tactile accessibility guidelines (\textbf{Q2}) was slightly higher for sourced graphics (\textbf{96.43\%}) compared to generated ones (\textbf{92.86\%}). Additionally, \textbf{28.57\%} of generated graphics required major edits, primarily due to excessive visual complexity, such as unintended 3D perspectives in objects like \textit{Chair} and \textit{Bed} (Figure~\ref{fig:main}), which can hinder tactile interpretability. A small fraction of sourced tactile images (\textbf{3.57\%}) were rejected, mostly due to human-induced mismatches in dataset curation—for example, incorrect natural image pairings, as seen in the camel species (Figure~\ref{fig:camel_mismatch}). This analysis highlights the strengths of AI-generated tactile graphics while emphasizing areas for improvement, particularly simplifying complex structures and refining dataset sourcing for higher semantic accuracy.


\subsection{Structural Fidelity and Abstraction Analysis}

To quantify how well the generated graphics ($G$) preserve visual and tactile-relevant structure, we computed Structural Similarity Index (SSIM) scores across three comparisons: between generated images and tactile benchmarks ($G$ vs. $T$), tactile benchmarks and natural images ($T$ vs. $N$), and between both generated/tactile graphics and natural image binary masks ($G$ vs. $N_{\text{bin}}$, $T$ vs. $N_{\text{bin}}$).

\begin{table}[t]
    \centering
    \caption{SSIM Comparison Across Modalities}
    \label{tab:ssim}
    \begin{tabular}{lcc}
        \toprule
        \textbf{Comparison} & \textbf{SSIM} & \textbf{Interpretation} \\
        \midrule
        $G$ vs. $T$ & 0.538 & Model matches human tactile design fidelity \\
        $T$ vs. $N$ & 0.549 & Human tactile abstraction baseline \\
        $G$ vs. $N_{\text{bin}}$ & 0.259 & Model preserves silhouettes \\
        $T$ vs. $N_{\text{bin}}$ & 0.215 & Human silhouette abstraction baseline \\
        \bottomrule
    \end{tabular}
\end{table}

As shown in Table~\ref{tab:ssim}, the similarity between $G$ and $T$ (0.538) nearly matches that of $T$ and $N$ (0.549), indicating that the model learns human-like abstraction patterns when generating tactile graphics. In silhouette comparisons using binary masks, $G$ outperformed $T$ (0.259 vs. 0.215), suggesting that the model retains object shape features more faithfully than traditional manual designs.

\begin{figure}[t]
    \centering
\includegraphics[width=0.99\columnwidth]{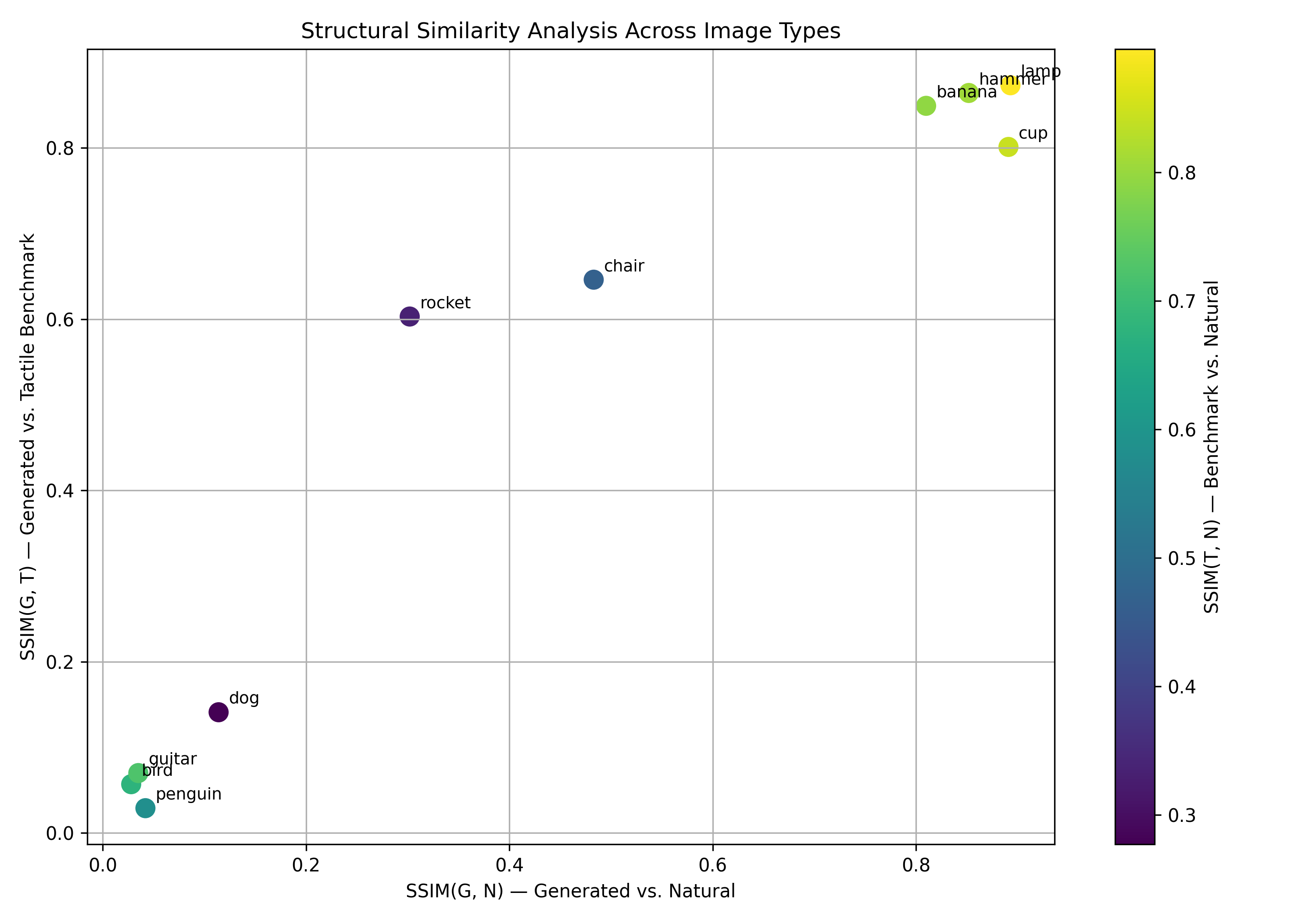}
    \caption{SSIM scores between generated ($G$), tactile benchmark ($T$), and natural ($N$) images across object classes. High-agreement classes (e.g., lamp, cup, banana) cluster in the upper right, while organic shapes (e.g., dog, bird) show greater divergence.}
    \label{fig:ssim_plot}
\end{figure}

Class-wise analysis (Figure~\ref{fig:ssim_plot}) highlights strong performance on structured objects like \textit{lamp}, \textit{cup}, and \textit{banana}, while more organic or cluttered shapes (e.g., \textit{dog}, \textit{bird}) showed lower alignment. Overall, the SSIM analysis confirms that the model not only replicates human tactile design patterns but often improves upon them in preserving essential structural features.


\subsection{Text-to-Image Translation Results}

\begin{figure}[ht]
\centering
\includegraphics[width=0.8\columnwidth]{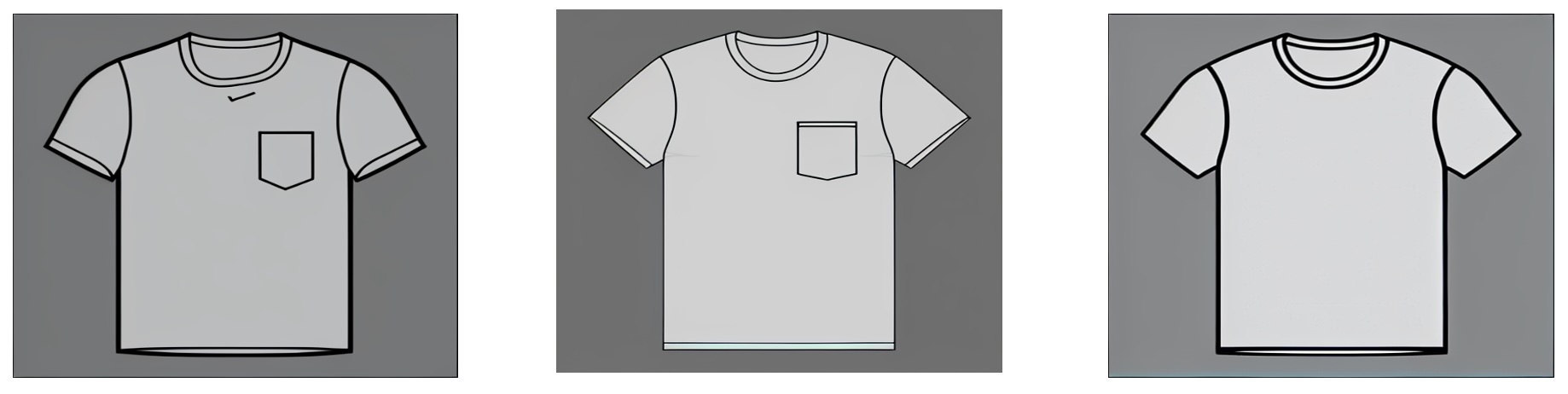} 
\caption{Prompt editing for customization: (Left) T-shirt generated using the base prompt; (Center) Logo removed by adding ``logo'' as a negative prompt; (Right) Pocket removed by omitting the keyword ``pocket'' from the prompt.}
\label{fig:shirt_edits}
\end{figure}

We generated a total of 32,000 tactile images across 66 classes, with 7,050 images retained after initial non-expert human filtering. Unlike the image-to-image translation setup, no natural image was provided as input; instead, the model relied solely on textual prompts to generate tactile graphics. Table~\ref{tab:data-summary} summarizes the per-class counts. Our fine-tuned adapters demonstrated adaptability through prompt-based edits (Figure~\ref{fig:shirt_edits}), enabling modifications such as logo removal or pocket additions. The prompt used for generating images was:  
\textit{Base Prompt: Create a tactile graphic of a t-shirt with a pocket for the visually impaired, highlighting the round neckline, pocket edges, and hem with raised lines. Ensure the pocket is a distinct raised rectangle, allowing users to discover and feel the detail on the shirt's chest.}

\section{Limitations and Future Work}
\label{sec:future_works}
\subsection{Limitations}
The current dataset is comprehensive but limited in size and diversity, with some classes having fewer images, which may hinder the model's generalization, while the reliance on class-specific adapters restricts scalability, highlighting the need for a more efficient, generalized model for practical applications.
\subsection{Future Work} 
Future efforts will focus on expanding TactileNet to include a diverse set of high-quality tactile images, enabling improved image-to-image translation systems that better manage complex tactile features. Incorporating generated images into the training pipeline will facilitate iterative model performance improvements, creating a self-reinforcing enhancement cycle. We aim to develop a single, generalized model capable of generating high-quality tactile graphics across all classes, leveraging our generated tactile images by employing techniques such as self-supervised training \cite{liu2021self,khan2022contrastive}. Continued collaboration with tactile graphics designers and educators will ensure the generated graphics align with real-world usability standards and incorporate expert feedback into the training and evaluation process.

\section{CONCLUSIONS}
\label{sec:conclusion}

This study marks a significant advancement in developing AI-driven accessibility tools through Generative AI, addressing the critical need for scalable, high-quality tactile graphics. Our key contributions are threefold:

\begin{itemize}
    \item \textbf{TactileNet Dataset}: A first-of-its-kind collection of 1,029 expert-curated tactile images across 66 classes, enabling AI model training aligned with tactile design principles.
    \item \textbf{Efficient Fine-Tuning}: Integration of LoRA and DreamBooth reduced labor costs while achieving 92.86\% adherence to tactile guidelines, rivalling manually sourced graphics.
    \item \textbf{Human-Centric Evaluation}: A protocol validated by tactile experts confirmed usability, with generated graphics requiring no rejections and achieving 100\% alignment with natural images.
\end{itemize}

By automating tactile graphic generation, our method accelerates production and democratizes access to educational materials for visually impaired learners. This work illustrates how AI can drive social good, offering a blueprint for inclusive technology that prioritizes human needs. \textbf{Our framework demonstrates how scalable, AI-augmented tools can empower tactile designers and educators, promoting inclusive learning environments worldwide.}

\section*{Acknowledgment}
This work was supported in part by MITACS and the Digital Alliance of Canada. We also thank the student volunteers at the Intelligent Machines Lab (iML), Carleton University, for their valuable contributions.










\bibliographystyle{ieeetr}
\bibliography{ref}

\end{document}